%% file: main.tex
\documentclass[sigconf]{acmart}

\usepackage{float}
\usepackage[utf8]{inputenc}
\usepackage{booktabs} 

\setcopyright{rightsretained}

\acmConference[E511'18]{Machine Learning for Signal Processing}{December 2018}{Bloomington, IN}
\acmYear{2018}
\copyrightyear{2018}

\begin{document}
\title{Yelp Food Identification via Image Feature Extraction and Classification}

\author{Fanbo Sun}
\affiliation{%
  \institution{Indiana University}
  \city{Bloomington}
  \state{Indiana}
}
\email{fanbsun@iu.edu}

\author{Zhixiang Gu}
\affiliation{%
  \institution{Indiana University}
  \city{Bloomington}
  \state{Indiana}
}
\email{zg3@iu.edu}

\author{Bo Feng}
\affiliation{%
  \institution{Indiana University}
  \city{Bloomington}
  \state{Indiana}
}
\email{fengbo@iu.edu}

\renewcommand{\shortauthors}{Sun et al.}

\begin{abstract}
Yelp has been one of the most popular local service search engine in US since 2004. It is powered by crowd-sourced text reviews and photo reviews. Restaurant customers and business owners upload photo images to Yelp, including reviewing or advertising either food, drinks, or inside and outside decorations. \footnote{Wikipedia: {https://en.wikipedia.org/wiki/Yelp}} It is obviously not so effective that labels for food photos rely on human editors, which is an issue should be addressed by innovative machine learning approaches. In this paper, we present a simple but effective approach which can identify up to ten kinds of food via raw photos from the challenge dataset. \footnote{Yelp Data Challenge: {https://www.yelp.com/dataset/challenge}}. We use 1) image pre-processing techniques, including filtering and image augmentation, 2) feature extraction via convolutional neural networks (CNN), and 3) three ways of classification algorithms. Then, we illustrate the classification accuracy by tuning parameters for augmentations, CNN, and classification. Our experimental results show this simple but effective approach to identify up to 10 food types from images.
\end{abstract}



\maketitle

\input{sections/introduction}

\input{sections/related-work}

\input{sections/methodology}

\input{sections/datasets-features}

\input{sections/experiments-results}

\input{sections/discuss-conclusion}

\bibliographystyle{ACM-Reference-Format}
\bibliography{sample-bibliography}

\end{document}

%% file: sections/introduction.tex
\section{Introduction}

Nowadays people really love taking photos, especially when they are in a fancy restaurant, in addition that smart mobile phones today are well equipped with high-resolution cameras. So, it is not a surprise that you can see there would be thousands of pictures from someone's phone album after a year. However, labeling and searching by words for these photos becomes a real hassle. For example, imaging you are talking with a friend about your extraordinary experience of eating a lobster in Bloomington IN, turning your phone upside down just want to share this picture, oops, you can not find it. This is because you forgot to label this picture and did not remember which day you took this picture.

Not just for personal photo album management, this issue raises astonishing importance for companies like Yelp, which does showing and researching business reviews. Digital photos identification poses a hard problem for those companies which rely on users' uploaded photos. Since they serve millions of customers and they may have billions of photos~\cite{YelpDataset2018}. It would be almost infeasible to edit photo labels by human editors. So, it would be promising that developing some automatically identification solution for user uploaded pictures.

More formally, in this work, we would like to build a model that can automatically classify a user uploaded food photo into a set of applicable categories, and the accuracy of prediction should be beyond average human guess. 
We first consider using pre-trained features provided by Yelp Data Challenge and feeding them to traditional machine learning algorithms like Convolutional Neural Network, Support Vector Machine, Gradient Boosting, with cross-validation. But unfortunately, the classification result was way below human eyeball. After double check, it becomes evident that the main issue lies on the features. As Yelp is a user-generated content platform, many pictures in our training set are vague, off-topic or mislabeled.

Given so, the decision was to do it from scratch. We carefully select 30 pictures for each class, use argumentation methods to enlarge our train set, manually extract features, finally the classification result is satisfactory.

The rest of the paper is organized as follows. Section~\ref{sec:related-work} discusses some related works, some of which either address similar issues or adopt similar solution methods that are related to the content of this work. Section~\ref{sec:methodology} presents the overview of our solution, machine learning models, algorithms, and et al. Section~\ref{sec:dataset} and ~\ref{sec:experiments-results} show the overview of the original datasets from Yelp and our experiments of showing the effectiveness of our implementation. Finally, this paper concludes the work and discusses some advantages and limitations of the work in Section~\ref{sec:dis-con}.

%% file: sections/related-work.tex
\section{Related Work}\label{sec:related-work}

In 2015, this work from Berkeley~\cite{Long_2015_CVPR} revealed that fully convolutional networks for semantic segmentation adopting AlexNet~\cite{NIPS2012_4824}, VGGNet~\cite{simonyan_very_2014}, and Google Net~\cite{szegedy_going_2015} can improve and simplify the state-of-art learning and inference. So, we also adopt the CNN architecture as a solution of image feature extraction, details of which can be found in Section~\ref{sec:methodology}.

Chan et al. proposes PCANet~\cite{chan_pcanet:_2015}, which is a simpel deep learning neural network for image classification for features relying on basic processing, such as PCA, binary hashing, and blockwise histogram. While our work does not rely on PCA, we use some image augmentation techniques which can improve accuracy.

While Convolutional Neural Network (CNN) is a promising approach to address many problems, Maggiori et al.~\cite{maggiori_convolutional_2017} presented an end-to-end framework, which adopts fully CNN approach to train large set of imperfect image sets with small sets of correctly labeled data sets. They demostrated this CNN architecture and its effective to classification problems. In this paper, our approach to address Yelp photo classification is similar to the extent where we mix large amount of low-confidence labeled datasets with augmentation datasets. We also use CNN for image feature extractions, in addition, we use image augmentation with xgboost and SVM classification methods comparing with direct CNN.

%% file: sections/methodology.tex
\section{Methodology}\label{sec:methodology}

\subsection{Overview}

\begin{figure*}
    \centering
    \includegraphics[width=0.9\textwidth]{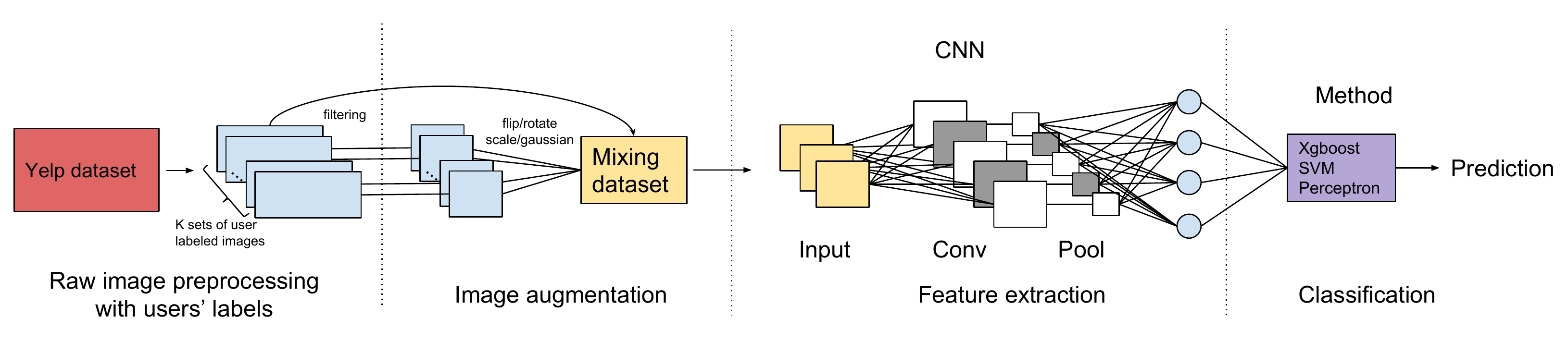}
    \caption{Architecture of Machine Learning Pipeline}
    \label{fig:architecture}
\end{figure*}

Figure~\ref{fig:architecture} presents the overall architecture of our machine learning pipeline. Our structure consists of four major steps: raw image preprocessing, image augmentation, feature extraction and classification. The raw Yelp dataset decomposes into K different clusters based on specific users' labels being chosen. Clean and correct images are filtered out manually from each cluster. After being applied different augmentation methods such as flip, rotate, scale, gaussian noise and GAN, these filtered images are mixed with both new augmented images and previous images from K clusters to form a big training dataset. A CNN is then used to extract features from the training dataset and finally we apply different classification methods such as Xgboost, SVM and MLP on those extracted features to get our classification accuracy.

\subsection{Image Preprocessing}

\subsubsection{Image Augmentation}
Given the Yelp photos are from users' uploads, most images are not directly usable to feed into Machine Learning Models. Because user uploaded images are usually taken from phone cameras with some distortion, which can largely degrade feature extraction and the final accuracy of classifications. After necessary transformations beforehand and filtering duplicates, the clean visible images seize below 10\% of the overall data sets. So, we use image augmentation techniques to enlarge our data sets in order to build accurate models. Some augmentation techniques used in our project are listed below:\par
1) Flip: Images are flipped both vertically and horizontally.\par
2) Rotate: We rotate images by 90, 180, and 270 degrees.\par
3) Scale: Because camera photos are usually take not far from objects, we scale images both outward and inward by only 10\%.\par
4) Gaussian Noise: We place random black and white pixels over images to simulate Gaussian noise.\par
5) Generative Adversarial Networks (not used but ideally works): The overall idea behind GANs is that you have two models playing with each other. As Ian Goodfellow describes it: one is a counterfeiter trying to produce seemingly real data while the other is a cop trying to determine what the fake counterfeit data is while trying to not raise false positives on real data, and in this way we can train neural networks to generate plausible data using a zero sum game. \par

More specifically, in this work, we apply the above techniques and generate 32 times of augmented images.

\begin{figure}
    \centering
    \includegraphics[width=0.45\textwidth]{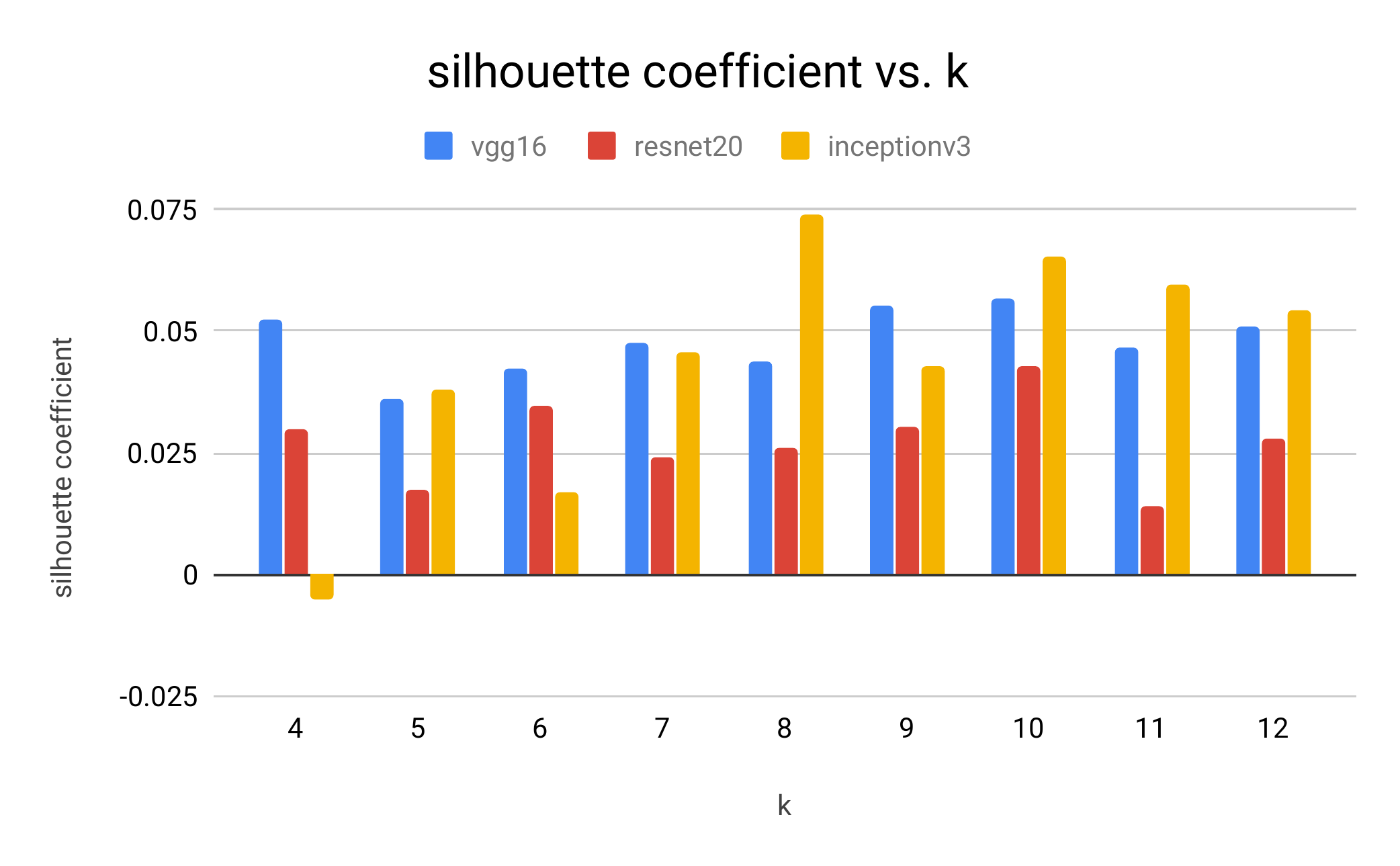}
    \caption{Sihouette coefficient vs. k in KMeans clustering}
    \label{fig:kmeans-validation}
\end{figure}

\subsection{Convolutional Neural Network}
Convolutional Neural Network (CNNs or ConvNets) are a category of feed-forward neural networks that are commonly used in areas such as image recognition and classification. They take advantage of spatial coherence between nearby elements from inputs, which allows them to have fewer weights as some parameters are shared. CNNs are composed of a number of convolutional and subsampling layers followed by pooling layers and fully connected layers. In this way, CNNs can easily transform the origin picture's pixel values into final scores with relatively low computation cost based on fewer parameters in the hidden layer.

\subsection{Image Feature Extraction}
Keras~\cite{Kerasapplication} as an open source library provides a set of state-of-the-art deep learning CNN models with pre-trained weights on ImageNet~\cite{imagenet}. These pre-trained models such as VGG16~\cite{simonyan_very_2014}, ResNet50~\cite{he2016deep} and InceptionV3~\cite{Szegedy_2016_CVPR} can be used for image feature extraction.

In order to show the effective of our prepared images for augmentation, we study Silhouette coefficient in KMeans clustering algorithm. 
Figure~\ref{fig:kmeans-validation} presents the Silhouette analysis results base on how many clusters can be selected from our prepared images for augmentation. In this figure, x-axis values are the number of clusters y-axis values are silhouette coefficient accordingly. The silhouette scores of VGG16 (blue bars) shows that using 10 clusters can obtain the highest scores, ResNet50 (red bars) shows also 10 clusters will get the highest scores and InceptionV3 (yellow bars) shows 8 will get the highest scores following by 10. The majority of models show their highest scores around 10 clusters , which means our ten types of food images are effective. 

Due to the reason that VGG16 has less weight layers and perform relatively faster and more stable around 10 clusters, we utilize VGG16 as a CNN model to extract features from our original datasets and new augmented datasets. We restrict the input size of images as 64X64 to boost our training speed and the final output of features has a dimension of 2048.

The Silhouette Coefficient used to valida $k$ clusters is defined as follow:
\[ \dfrac{b-a}{max(a, b)} \], in which the Silhouette Coefficient is calculated using the mean intra-cluster distance ($a$) and the mean nearest-cluster distance ($b$) for each sample.~\cite{sklearn}

\subsection{Image Classification}

\subsubsection{Xgboost}
Gradient Boosting
We tried gradient boost for this problem, specifically we used XGBoost, which is an implementation of gradient boosted decision trees designed for speed and performance.~\cite{chen_xgboost:_2016}

\subsubsection{Support Vector Machine (SVM)}
Support Vector Machine (SVM) is a supervised machine learning algorithm which can be used for both classification or regression challenges. we plot each data item as a point in n-dimensional space, where n is number of features you have, with the value of each feature being the value of a particular coordinate. Then, we perform classification by finding the hyper-plane that differentiate the two classes very well.

We applied SVM with a linear kernel and RBF kernel on the data set. In order to get better performance, we also use grid search method to tune the C value.

\subsubsection{Radial Basis Function kernel}
Radial Basis Function kernel (RBF) is a kernel function which extends to patterns that are not linearly separable by transformations of original data to tap into new space, defined as follow:
\[ K(x, x^\prime) = exp(- \frac{\left\|x-x\prime\right\|^2}{2\sigma^2}) \]

\subsubsection{Multilayer Perceptron}
Neural network is a series of algorithms that endeavors to recognize underlying relationships in a set of data through a process that mimics the way the human brain operates. Neural networks can adapt to changing input so the network generates the best possible result without needing to redesign the output criteria.

\subsubsection{Cross Validation}
Considering there is no ground truth dataset for testing, we use the training set splits for both training and testing. In our three classification methods, all tests are tuned to take 70\% as training sets and 30\% as testing sets.

%% file: sections/datasets-features.tex
\section{Datasets and Features}\label{sec:dataset}

\begin{figure}[ht]
    \centering
    \includegraphics[width=0.35\textwidth]{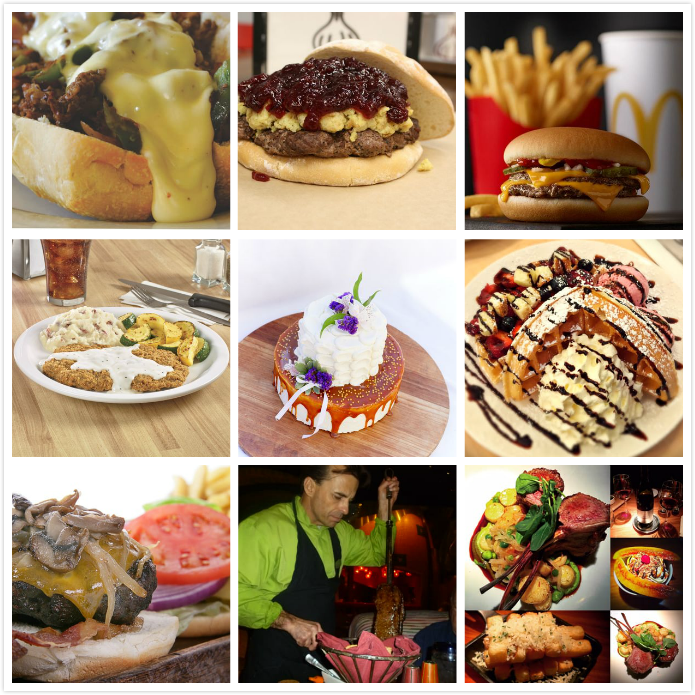}
    \caption{Representative sample photos from Yelp}
    \label{fig:data_overview}
\end{figure}

Yelp publishes over 280,000 pictures from over 2000 businesses, the dataset of which can be found on the Yelp Data Challenge Website~\cite{YelpDataset2018}. This dataset consists of inside, outside, drink, and food photos. Yelp has labeled the above four categories, while subcategories for specific types of food are missing. Most photos are uploaded either by customers or business owners. These photos may have valid captions, which are correctly descriptions to corresponding photos. There are many photos which do not have any captions or have incorrect captions. For example, a restaurant which sells both burgers and sandwiches tends to write both "burger" and "sandwich" for burger photos and sandwich photos for better marketing. In summary, there is no ground truth labeled image sets. Figure~\ref{fig:data_overview} shows a few representative sample photos. Most food images contain noisy background, including tables, plates, chefs, and etc. Some food images contain more than two kinds of food. So, the baseline accuracy of classification is lower than 40\% because of in-perfect word captions from users.

To isolate the scope of this work and increase the accuracy, initially, we select photos based on observations and keywords from rural user captions.
According to our observations, the top ten most popular types are covered:  1) Burger, 2) Cake, 3) Chips, 4) Noodles, 5) Pizza, 6) Roll, 7) Salad, 8) Sandwich, 9) Steak, and 10) Wings. These ten categories are, comparatively, simpler and clearer to identify, so in this work we use them for the classification task.

%% file: sections/experiments-results.tex
\section{Experiments and Results}\label{sec:experiments-results}

\subsection{Evaluation Metrics}
The evaluation metrics for our experiment is the prediction accuracy. This accuracy comes from the fraction of data for which the classification algorithm correctly predicted if it can be labeled with the attribute or not. Specifically, for our photo dataset, we would be analyzing the accuracy for all the validation sets of raw k clusters' dataset, augmented dataset and mixing dataset.
\subsection{Experiments}

\begin{table}[ht]
\setlength{\arrayrulewidth}{1mm}
\renewcommand{\arraystretch}{1.5}
\begin{tabular}{lp{.6in}p{.6in}p{.6in}}
\toprule
Category & Multilayer  Perceptron & Gradient  Boosting& Support Vector Machine \\
\hline
Original        & 28.16\%              & 43.20\%              & 36.13\%   \\
Augmentation    & 82.38\%              & 79.96\%              & 90.19\%         \\
Mixed           & 53.40\%              & 65.82\%              & 68.49\%        \\ \bottomrule
\end{tabular}
\caption{Classification Result}
\label{table:classresult}
\end{table}

After manually selecting the images with correct labels, we come up with a very small but relatively clean dataset with 10 selected labels. Each of cluster contains 30 images. Due to the ineffectiveness of small training data, we apply image augmentation to every image and enhance one image into 32 augmented images. Therefore, the size of clean dataset increases from $10\times30$ to $10\times30\times32$. For the image feature extraction, we used VGG16 as a CNN model to extract image to the dimension of 2048.

For each method of classification, we utilized toolboxes for a simple and quick test. In SVM method, we chose RBF kernel with the relative parameter to do the non-linear transformation to the dataset. For MLP method, we created a five layer nerual network with two dropout layer and three activation layer. Considering the output layer result should be in the range of 0 to 10, we selected the activation function of final layer to be RELU. The first and the second activation layer are with RELU and sigmoid function. Between each activation layer, a drop out layer is applied to prevent overfitting and gradient boosting.  

All the experiments except SVM were trained on GeForce GTX 960 GPUs. MLP took about 10 minutes to train 1000 epochs on each dataset and XGboost also took not more than 10 minutes. However, because our toolbox for SVM did not supply GPU computing, instead, we ran SVM on school's machine with 64 cores CPU. It took around 2 hours to train the raw dataset and over 6 hours to train the mixing dataset. 
\subsection{Results}

\subsubsection{Image Feature Extraction}
The augmentation set has a total dimension of $9280\times2048$, the raw training set has a dimension of $4829\times2048$ and the mixing set has a dimension of $14109\times2048$. After doing cross validation to each set, we got our testing sets which are in the dimension of $1856\times2048$ for aug, $966\times2048$ for raw and $2822\times2048$ for mix respectively.

\subsubsection{Image Classification}
In order to avoid overfitting, we perform cross validation. It’s very similar to train/test split, but creating more subsets. Meaning, we split our data into k subsets, and train on k-1 one of those subset. What we do is to hold the last subset for test. We perform such operation for each of the subsets.

Table~\ref{table:classresult} shows the testing accuracy of each classification method. 

Clearly, before augmentation, our classification algorithms perform poorly and even way below human eyeball. While in our mixed data set, the Support Vector Machine achieves accuracy around 68.49$\%$, pretty closes to our expectation.

%% file: sections/discuss-conclusion.tex
\section{Discussions, Conclusions and Future Work}\label{sec:dis-con}

As we tested k cluster validation with KMeans algorithms, the range of k varies from 4 to 12 by using three Image Model via transfer learning: VGG16, resnet20, and inceptionv3. All of them show effectiveness on our training sets. However, due the limitation of time and computing resources, we can only extract features with VGG16, even though incpetionv3 may produce better features.

Even after manually select small sample images from the datasets and augmentation, there are still no ground truth datasets for training. This is mainly due to user uploaded pictures can be very vague and business owners tend to add more irrelevant captions for pictures as many as possible. All these facts can result in a less accurate design of our CNN for features extraction and classification, in which pooling and dropout layers may be not well studied.

The main contribution of this paper is, given low confidence about the quality of the training datasets, image augmentation shows its effectiveness, comparing the classification accuracy of before applying this technique and after. The training accuracy for 10 types of food can be as low as 28\% in a multi-layer perceptron. However, after augmentation, the accuracy of classification can be as high as around 70\%. 

Furthermore, it would be interesting to see some more novelty augmentation methods in application, for example, "style transfer" can be used to augment data in situations where the available data set is unbalanced. Also, business-related metrics such as user comments, the numbers of likes, are supposed to be beneficial to the increasing the classification accuracy.

